\documentclass[sigconf, authorversion]{acmart}
\usepackage{xcolor}
\definecolor{urlcolor}{RGB}{251, 49, 153}

\AtBeginDocument{%
  }

\copyrightyear{2022} 
\acmYear{2022} 
\setcopyright{acmcopyright}
\acmConference[MM '22]{Proceedings of the 30th ACM International Conference on Multimedia}{October 10--14, 2022}{Lisboa, Portugal}
\acmBooktitle{Proceedings of the 30th ACM International Conference on Multimedia (MM '22), October 10--14, 2022, Lisboa, Portugal}
\acmPrice{15.00}
\acmDOI{10.1145/3503161.3547999}
\acmISBN{978-1-4503-9203-7/22/10}

\begin{document}

\title{Progressive Limb-Aware Virtual Try-On}
\author{Xiaoyu Han}
\affiliation{%
  \institution{Harbin Institute of Technology}
  \city{Weihai}
  \country{China}
}
\email{xyhan@stu.hit.edu.cn}

\author{Shengping Zhang}
\affiliation{%
  \institution{Harbin Institute of Technology}
  \city{Weihai}  
  \country{China}
}
\email{s.zhang@hit.edu.cn}
\authornote{Corresponding author.}

\author{Qinglin Liu}
\affiliation{%
  \institution{Harbin Institute of Technology}
  \city{Weihai}  
  \country{China}
}
\email{qinglin.liu@outlook.com}

\author{Zonglin Li}
\affiliation{%
  \institution{Harbin Institute of Technology}
  \city{Weihai}
  \country{China}
  }
\email{zonglin.li@hit.edu.cn}

\author{Chenyang Wang}
\affiliation{%
  \institution{Harbin Institute of Technology}
  \city{Weihai}
  \country{China}
  }
\email{c.wang@stu.hit.edu.cn}

\renewcommand{\shortauthors}{Xiaoyu Han et al.}


\begin{abstract}
Existing image-based virtual try-on methods directly transfer specific clothing to a human image without utilizing clothing attributes to refine the transferred clothing geometry and textures, which causes incomplete and blurred clothing appearances.
In addition, these methods usually mask the limb textures of the input for the clothing-agnostic person representation, which results in inaccurate predictions for human limb regions (i.e., the exposed arm skin), especially when transforming between long-sleeved and short-sleeved garments.
To address these problems, we present a progressive virtual try-on framework, named PL-VTON, which performs pixel-level clothing warping based on multiple attributes of clothing and embeds explicit limb-aware features to generate photo-realistic try-on results.
Specifically, we design a Multi-attribute Clothing Warping (MCW) module that adopts a two-stage alignment strategy based on multiple attributes to progressively estimate pixel-level clothing displacements.
A Human Parsing Estimator (HPE) is then introduced to semantically divide the person into various regions, which provides structural constraints on the human body and therefore alleviates texture bleeding between clothing and limb regions.
Finally, we propose a Limb-aware Texture Fusion (LTF) module to estimate high-quality details in limb regions by fusing textures of the clothing and the human body with the guidance of explicit limb-aware features.
Extensive experiments demonstrate that our proposed method outperforms the state-of-the-art virtual try-on methods both qualitatively and quantitatively.
The code is available at \textcolor{urlcolor}{\url{https://github.com/xyhanHIT/PL-VTON}}.
\end{abstract}

\begin{teaserfigure}
  \includegraphics[width=\textwidth]{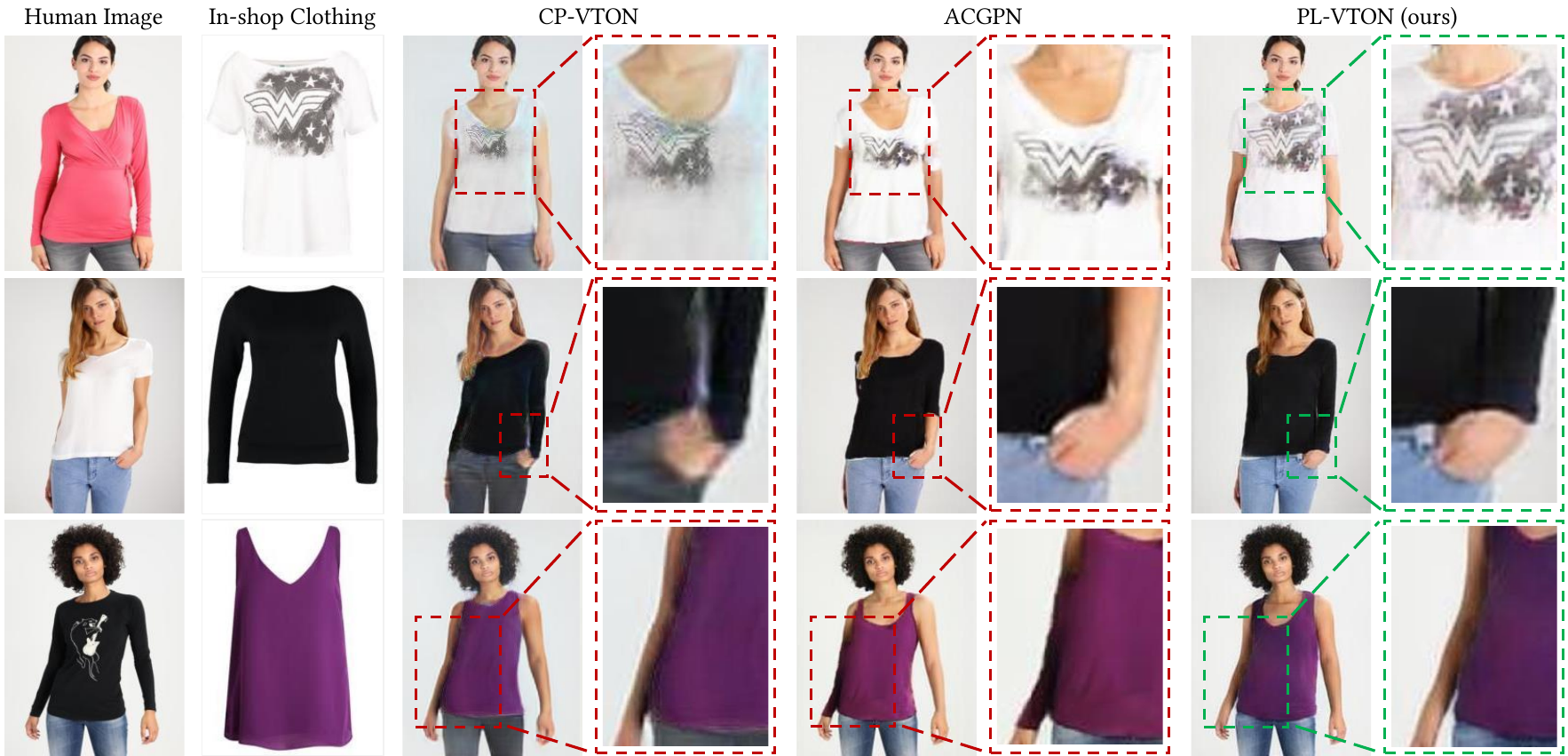}
  \caption{Given the human images and in-shop clothing images, our proposed PL-VTON produces high-quality try-on results. In particular, PL-VTON outperforms the state-of-the-art methods in scenes with the complex clothing textures (the first row) and the transformation between long-sleeved and short-sleeved clothing (the second row and the third row).}
  \label{fig:teaser}
\end{teaserfigure}


%

\maketitle

\section{Introduction}
Virtual try-on has attracted a lot of attention in recent years because of its wide applications in e-commerce and fashion image editing, which aims to transfer specific clothing to a human image.
Existing methods can be roughly classified into 3D-based methods~\cite{guan2012drape, yang2016detailed, pons2017clothcap, chen2016synthesizing} and 2D-based methods~\cite{han2018viton, wang2018toward, han2019clothflow, jandial2020sievenet, jetchev2017conditional, chopra2021zflow, yang2021ct, li2021toward}. 
3D-based methods utilize computer graphics for building 3D models and obtain the result through complex rendering, which has excellent control over the material and the clothing deformation. However, these methods rely heavily on complex data representations and consume massive computational resources. 
Conversely, 2D-based methods are more suitable for real-world scenarios due to the lightweight data.

The early 2D-based method CA-GAN~\cite{jetchev2017conditional} leverages the generative adversarial network (GAN) to directly generate try-on images without other descriptive person representations, which fails to get photo-realistic results. 
VITON~\cite{han2018viton} first proposes the idea of using the thin-plate spline (TPS) transformation to perform clothing warping, which greatly improves the try-on performance.
Then CP-VTON~\cite{wang2018toward} adopts an improved TPS transformation with full learnable parameters via a deep convolutional neural network to obtain a more robust result of the clothing alignment. 
However, each TPS parameter controls the deformation of a coarse pixel block, so these TPS-based methods are limited by the low degrees of freedom, which produce distorted and incorrect results when large geometric deformations occur on the clothing.
ClothFlow~\cite{han2019clothflow} replaces TPS with the dense pixel-wise appearance flow, but the high degrees of freedom and the lack of proper regularization cause unstable pixel displacements and lead to texture bleeding in the try-on result.
We argue that when the estimation of the clothing deformation is constrained by a single stage, the network needs to allocate additional attention to aligning clothing with the human body from other attributes besides the clothing geometry (e.g., the adjustment in spatial location and size), which reduces the finesse and controllability of the clothing warping.

On the other hand, it is difficult to obtain the data of a person wearing different clothing in a fixed pose, so recent methods usually construct the clothing-agnostic person representation to eliminate the effects of the source clothing item and train the network in a self-supervised way. 
Some of them~\cite{yu2019vtnfp, dong2019towards, chen2021fashionmirror} mask the clothing regions in the human image along the clothing boundary. 
However, the rough outline of the clothing is still preserved, which does not keep clothing-agnostic strictly.
Others~\cite{wang2018toward, minar2020cp, han2018viton, xie2021towards, jandial2020sievenet} mask the clothing regions and limb regions (i.e., the exposed arm skin in the human image), which loses limb texture information and is negative to the texture retention.

In this paper, we propose PL-VTON to perform stable clothing warping and retain realistic limb textures in the try-on result (as shown in Figure \ref{fig:teaser}), which consists of three progressive modules: (1) Multi-attribute Clothing Warping (MCW), which adopts a two-stage alignment strategy to align the clothing with the human body. In-shop clothing is first adapted in spatial location and size according to the explicit semantic parsing of the paired human image, then a multi-scale flow predictor is used to estimate the geometric deformation of the clothing precisely. (2) Human Parsing Estimator (HPE), which aims to produce the prior guidance for the limb texture extraction and provide structural constraints for the subsequent generation of the try-on result by predicting the parsing map of the person wearing the target clothing. (3) Limb-aware Texture Fusion (LTF), where a coarse-to-fine texture fusion scheme with the limb-aware guidance is applied to preserve high-quality limb textures in the final try-on result.

The contributions of this paper are summarized below:
\begin{itemize}
    \item
    We propose a virtual try-on framework named PL-VTON, which generates high-quality try-on results progressively by multi-attribute clothing warping, parsing map estimation, and texture fusion with the limb-aware guidance. 
    \item
    We design a novel multi-attribute clothing warping module that utilizes a two-stage alignment strategy based on multi-attribute to obtain the fine-grained clothing deformation and warp the clothing precisely.
    \item
    We present a novel limb-aware texture fusion module to fuse the textures of the clothing and the human body from coarse to fine, where the explicit limb-aware guidance has a positive impact on the retention of limb details.
    \item
    Extensive experiments on the VITON dataset demonstrate the significant superior performance in the image-based virtual try-on task achieved by our PL-VTON.
\end{itemize}

\section{Related Work}
\subsection{Flow Estimation}
Flow estimation aims to find the pixel-level correspondence between two frames or two images. Specifically, it describes which pixels in the source can be used to synthesize the target and how those pixels are transferred to the specific positions in the result.
Early flow estimation is usually used in video tasks, which is called optical flow estimation. The optical flow estimation attempts to learn a siamese network by taking two consecutive video frames as the input, then the raw pixels of the first frame are warped to the next one.
FlowNet~\cite{dosovitskiy2015flownet} is the first CNN-based end-to-end version for the flow estimation, which uses a U-net architecture to predict the optical flow directly.
FlowNet2~\cite{ilg2017flownet} adopts the stacked hourglass network with more training data and more complex training strategies to improve the accuracy in small motion areas.
PWC-Net~\cite{sun2018pwc} utilizes well-established principles of pyramidal processing, warping, and cost volume processing to predict the optical flow, which further promotes the performance and reduces the model size simultaneously.
RAFT~\cite{teed2020raft} proposes a novel deep network architecture with a recurrent unit, which uses many lightweight and recurrent update operators to estimate the optical flow.

In addition, flow estimation is also applied to predict a 2D vector field without timing information, which warps the source image to the target image based on the similarity in appearance. It is defined as the appearance flow by Zhou et al.~\cite{zhou2016view}.
Appearance flow has been widely applied in the field of computer vision. For instance, StructureFlow~\cite{ren2019structureflow} adopts appearance flow in image inpainting. To generate realistic alternative contents for missing holes, the offset vectors are predicted to flow the pixels from source regions to missing regions.
Document image rectification~\cite{ma2018docunet, das2019dewarpnet, li2019document, markovitz2020can, feng2021docscanner, feng2021doctr} adopts the appearance flow network to regress a dense 2D vector field that samples pixels from the distorted document images to the rectified ones.
Appearance flow is also used in human pose transfer, e.g., Li et al.~\cite{li2019dense} fit a 3D model to the given pose pair and project them back to the 2D plane to compute the dense appearance flow. With the pixel-level displacements, feature warping is performed on the human image and the photo-realistic result of the target pose is generated.
In this paper, we adopt the appearance flow to represent the clothing deformation, which acts as the prior information for generating the try-on result.

\subsection{Fashion Analysis and Virtual Try-on}
Fashion analysis has attracted considerable attention in recent years due to its wide range of applications. The initial focus is more on the classical tasks such as clothing classification~\cite{bossard2012apparel, chen2012describing}, clothing recognition~\cite{yang2011real, kalantidis2013getting}, clothing segmentation~\cite{yang2014clothing, wang2011blocks, ji2018semantic}, and 
fashion image retrieval~\cite{corbiere2017leveraging, gajic2018cross, lang2020plagiarism, d2021localized}. Recently, more novel tasks such as fashion style estimation~\cite{kiapour2014hipster, veit2015learning, simo2016fashion, hsiao2017learning}, clothing recommendation~\cite{han2017learning, liu2016deepfashion, sha2016approach, ma2017towards, yu2018aesthetic, hidayati2018dress}, popularity predicting~\cite{yamaguchi2014chic}, and fashion landmark localization~\cite{ge2019deepfashion2, lee2019global, liu2016fashion, yan2017unconstrained, wang2018attentive} have been explored, where virtual try-on is one of the most challenging visual tasks because it requires both capturing the clothing accurately and merging clothing textures properly with the given human body.

Existing methods for virtual try-on are mainly based on 3D modeling~\cite{guan2012drape, yang2016detailed, pons2017clothcap, chen2016synthesizing} or 2D images~\cite{han2018viton, wang2018toward, han2019clothflow, jandial2020sievenet, jetchev2017conditional, chopra2021zflow, yang2021ct, li2021toward}.
3D-based methods rely on sophisticated instruments and higher-dimensional calculations to achieve try-on effects, but the complicated process constrains the application value. Conversely, 2D-based methods are more broadly applicable.
CA-GAN~\cite{jetchev2017conditional} applies the generative adversarial network (GAN) to learn the relation between the in-shop clothing image and the human image. Based on the relation, the image of the person wearing new clothing is then generalized and generated.
VITON~\cite{han2018viton} proposes a virtual try-on network without using 3D information in any form. It uses thin-plate spline (TPS) transformation to warp the clothing and an encoder-decoder is then used to generate the try-on result. Besides, it adopts a clothing-agnostic person representation to eliminate the effects of the original clothing item in the human image.
To improve the accuracy of the warping process, CP-VTON~\cite{wang2018toward} proposes to utilize a deep convolutional neural network to dynamically estimate the TPS parameters.
CP-VTON+~\cite{minar2020cp} is designed based on the pipeline structure of CP-VTON, which corrects the erroneous clothing-agnostic person representation in the dataset and makes the network input more reasonable.
SieveNet~\cite{jandial2020sievenet} further optimizes the clothing warping, where a coarse-to-fine clothing warping network is trained with a novel perceptual geometric matching loss to better model fine intricacies while transforming the target clothing to align with the human body.

However, the above TPS-based methods cannot handle large geometric deformation well due to the limited degrees of freedom. To warp the clothing more freely, some recent methods adopt the appearance flow in their framework.
ClothFlow~\cite{han2019clothflow} proposes a generative model based on the appearance flow to improve the freedom of clothing deformation. By estimating the dense pixel-level flow between the source and target clothing regions, the model can effectively simulate the geometrical changes and generate the warped clothing image.
Nonetheless, due to the unrestricted degrees of deformation and lack of proper regularization, this scheme may lead to drastic and unrealistic clothing warping and cause artifacts or texture bleeding in the final try-on result.
Zflow~\cite{chopra2021zflow} points out the shortcomings of~\cite{han2019clothflow} and adopts a gated aggregation of hierarchical flow estimates and some geometric priors as the improvement.
PF-AFN~\cite{ge2021parser} proposes a “teacher-tutor-student” knowledge distillation strategy and formulates it as distilling the appearance flows between the clothing image and the human image, which aims to find accurate dense correspondences between them to produce high-quality results.
Inspired by the above methods, we use an appearance flow that considers multiple scales to represent the clothing deformation, which is produced by a two-stage alignment strategy progressively. In addition, we adopt the limb-aware guidance to optimize the clothing-agnostic person representation, assisting the network in generating high-quality limb details.

\begin{figure*}[!t]
    \centering
    \includegraphics[width=1\textwidth]{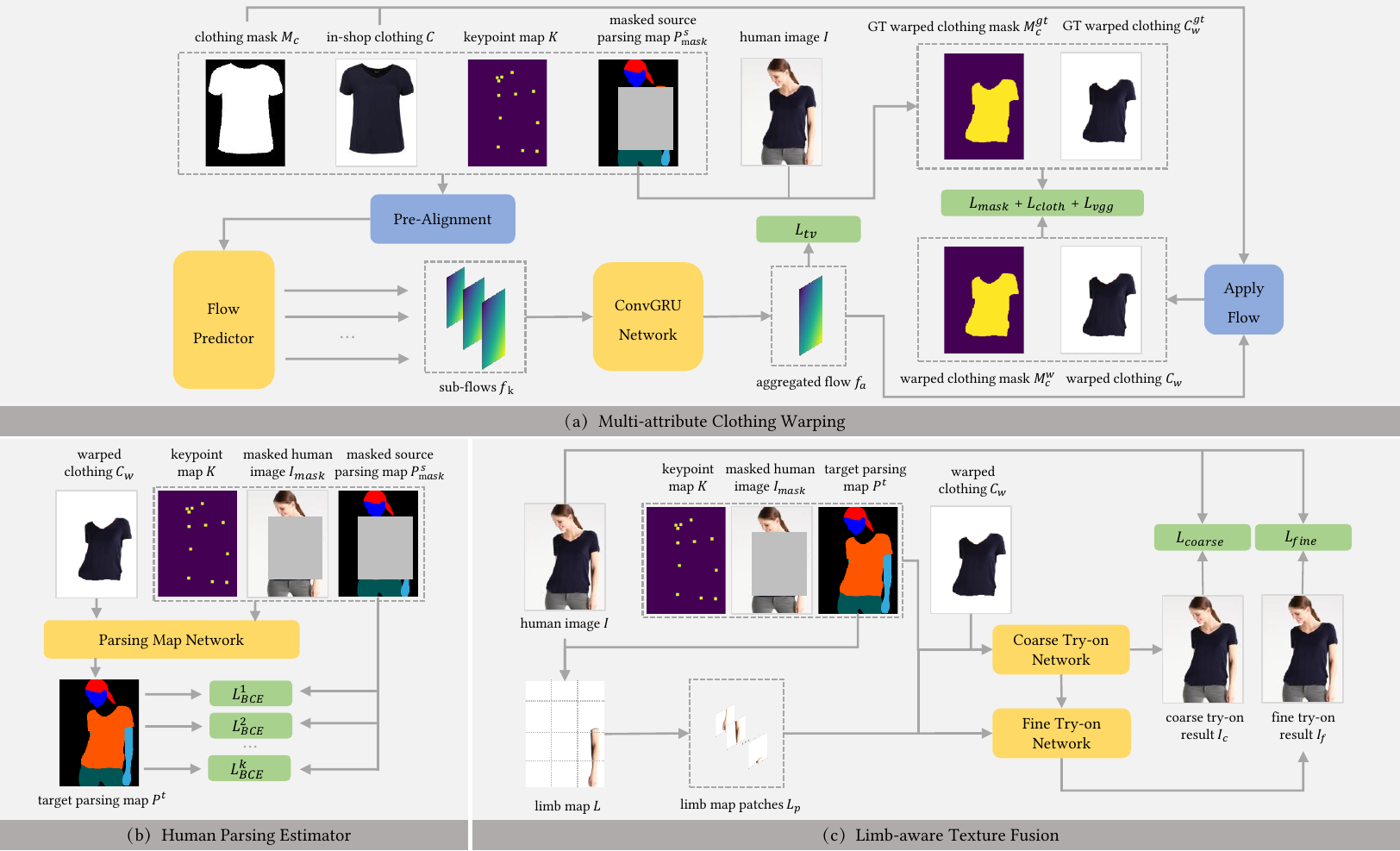}
    \caption{The overview of the proposed PL-VTON. (a) MCW adopts a two-stage alignment strategy to estimate the aggregated flow $f_a$. (b) HPE estimates the target parsing map $P^t$ to provide structural constraints. (c) LTF first produces the coarse try-on result $I_c$ and then utilizes the information of the limb map to refine $I_c$ and get the fine try-on result $I_f$.}
    \label{fig:framework} 
\end{figure*}

\section{METHOD}

\subsection{Overview}
The overview of our proposed progressive limb-aware virtual try-on framework (PL-VTON) is shown in Figure \ref{fig:framework}, which consists of three modules: Multi-attribute Clothing Warping (MCW), Human Parsing Estimator (HPE), and Limb-aware Texture Fusion (LTF). These modules are cascaded to generate the high-quality virtual try-on result progressively.
Specifically, given an in-shop clothing $C$, an 18-channel keypoint map $K$, and a 7-channel source parsing map $P^s$, MCW aims to estimate an aggregated flow $f_a$, which is applied on $C$ to get a warped clothing $C_w$.
Then HPE uses $C_w$, $K$, $P^s$, and a masked human image $I_{mask}$ to predict a target parsing map $P^t$, which represents the semantic segmentation of the person wearing the target clothing.
Finally, LTF takes $C_w$, $K$, $P^t$, and $I_{mask}$ as the input to first perform the coarse texture fusion and output a coarse try-on result $I_c$, then a set of limb map patches $L_p$ (obtained through $P^t$ and the human image $I$) is adopted to refine $I_c$ and get a fine try-on result $I_f$.

\subsection{Multi-attribute Clothing Warping}
As illustrated in Figure \ref{fig:framework} (a), the proposed Multi-attribute Clothing Warping (MCW) adopts a two-stage alignment strategy: the spatial location and size of the clothing are first adjusted to roughly align the clothing with the human body according to an adaptive circumscribed rectangle, then a multi-scale flow predictor is used to estimate the geometric deformation of the clothing precisely and get the warped clothing $C_w$.

\subsubsection{\textbf{Clothing-agnostic Input}} 
\begin{figure}[!t]
  \centering
  \resizebox{\linewidth}{!} {
    \includegraphics{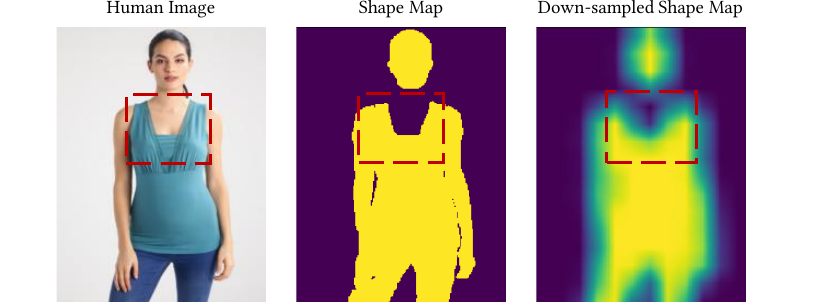}
  }
  \caption{The human shape map contains geometric information of the clothing (e.g., the original collar shape), even if it has been down-sampled and interpolated.}
  \label{figure: shape}
\end{figure}

To train the virtual try-on network, a straightforward approach is to leverage the training data of the person wearing different clothing in a fixed pose, which is usually difficult to acquire.
It is a good practice to obtain the clothing-agnostic person representation~\cite{han2018viton} of a human image (i.e., the clothing regions in the human image are masked before entering the network) and send it to the network, so we can treat the human image as both ground truth and input.
We apply this representation and further make some improvements for obtaining the prior. We design an adaptive mask scheme, where a circumscribed rectangle of the clothing regions in $I$ is used as the mask (obtained by the extreme values of the clothing pixel position in four directions), and it is modified according to the limb regions (i.e., the exposed arm skin in the human image is added to the masked regions). This mask is used to occlude the source parsing map $P^s$ and the human image $I$, which ensures that the input does not contain the original clothing information.
Note that we do not adopt the human shape map like~\cite{han2018viton, minar2020cp, jandial2020sievenet}, since it contains the geometric information of the clothing (e.g., the original collar shape) and is not a clothing-agnostic representation strictly, even if it has been down-sampled and interpolated, which is shown in Figure \ref{figure: shape}.

\subsubsection{\textbf{Pre-alignment in Spatial Location and Size}} 
Generally, the in-shop clothing $C$ and the clothing in $I$ are different in spatial location and size. It is necessary to pre-align the clothing in these two attributes before estimating the appearance flow, which helps the network to handle fine-grained geometric deformation, and the whole warping process becomes more generalizable and controllable because of the separate two-stage alignment strategy.

Therefore, we design an adaptive way to align the in-shop clothing $C$ with the human image $I$ in spatial location and size before geometric deformation. 
First, for the in-shop clothing mask $M_c$, we get the center point of its circumscribed rectangle $(x_1, y_1)$. Similarly, for the clothing regions of $I$, we get the center point of its circumscribed rectangle $(x_2, y_2)$. Then a shift operation is performed on the in-shop clothing $C$ with the movement vector $(x_1-x_2, y_1-y_2)$ to get the shifted clothing image $C_l$.
Next, the height of the clothing regions in $C$ and $I$ are calculated as $h_s$ and $h_t$, respectively, and we get a size ratio as $h_s/h_t$. Note that we do not use the width because some complex human postures seriously affect the horizontal value of the clothing regions in $I$, which cannot reflect the real size. We use the size ratio $h_s/h_t$ to resize the shifted clothing image $C_l$ and get the resized clothing image $C_s$.

\subsubsection{\textbf{Multi-scale Flow Predictor}}
The multi-scale flow predictor aims to produce a set of sub-flows from different scales and aggregate them into a final precise appearance flow to indicate the geometric deformation of clothing.
The backbone network is a ResNet34~\cite{he2016deep} with a 5-layer decoder, and its inputs are the resized clothing image $C_s \in \mathbb{R}^{3 \times H\times W}$, the keypoint map $K \in \mathbb{R}^{18 \times H\times W}$ and the source parsing map $P^s \in \mathbb{R}^{7 \times H \times W}$. The $k$-th layer of the decoder is used to predict the sub-flow $f_k \in \mathbb{R}^{ 2 \times \frac{H}{6-k} \times \frac{W}{6-k}}$, where $k\in\mathbb\{1,2,...,5\}$.
Inspired by~\cite{chopra2021zflow}, we adopt a convolution gated recurrent unit (ConvGRU)~\cite{siam2017convolutional} to aggregate multi-scale sub-flows into a final precise flow, which uses updating and resetting operations to gate these sub-flows and combines them through multiple non-linear weighted summations for the fine-grained deformation.

\subsubsection{\textbf{Loss}} 
We apply the aggregated flow $f_a$ on the in-shop clothing mask $M_c$ to obtain the warped clothing mask $M^w_c$. This result is used to calculate the mask loss $L_{mask}$ to constrain the clothing geometry, which is formulated as:
\begin{equation}
    L_{mask} = \Vert M^{w}_c - M^{gt}_c \Vert_1
\end{equation}
where $M^{gt}_c$ is extracted from the source parsing map $P^s$, it represents the clothing regions in the human image $I$.
Furthermore, $M^{gt}_c$ is also applied to get the clothing $C^{gt}_w$ in the human image, which is regarded as the ground truth of the warped clothing $C_w$:
\begin{equation}
   C^{gt}_w = I \odot M^{gt}_c
\end{equation}
where $\odot$ is element-wise multiplication. The clothing loss $L_{cloth}$ and the perceptual loss $L_{vgg}$ are formulated as:
\begin{equation}
    L_{cloth} = \Vert C_{w} - C^{gt}_w\Vert_1
\end{equation}
\begin{equation}
    L_{vgg} = \sum^n_{i=1} \lambda_i \Vert \phi_i(C_{w}) - \phi_i(C^{gt}_w)\Vert_1
\end{equation}
where $\phi_i(\cdot)$ denotes the feature maps of the i-th layer in the visual perception network VGG19~\cite{simonyan2014very} pre-trained on ImageNet~\cite{deng2009imagenet}, and $\lambda_i$ is the corresponding weight.
In addition, a total variation loss~\cite{fan2018end} is applied to ensure the spatial smoothness for the flow $f_a$:
\begin{equation}
    L_{tv} =\sum_{i \in \Omega}\sqrt{{\rm D_x}(f_a^i)^2+{\rm D_y}(f_a^i)^2}
\end{equation}
where ${\rm D_x}(\cdot)$ and ${\rm D_y}(\cdot)$ are the horizontal and vertical difference functions, respectively. $\Omega$ is the area of $f_a$.
Finally, the whole objective function of MCW is presented as:
\begin{equation}
    L_{MCW} = \lambda_{mask} L_{mask} + \lambda_{cloth} L_{cloth} + \lambda_{vgg} L_{vgg} + \lambda_{tv} L_{tv}
\end{equation}

\begin{figure*}[!t]
  \centering
  \resizebox{\linewidth}{!} {
    \includegraphics{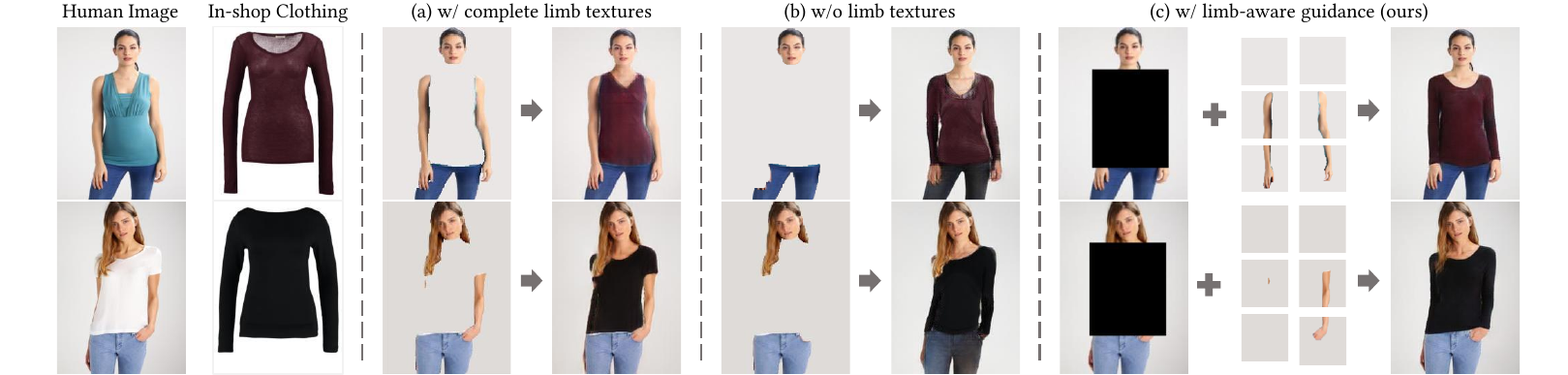}
  }
  \caption{The effect of the amount of limb information used to produce the try-on result.}
  \label{figure:limb}
\end{figure*}

\subsection{Human Parsing Estimator}
Due to the neglect of structural information, some existing methods~\cite{han2018viton, wang2018toward, minar2020cp} suffer from texture bleeding between clothing and human skin, which severely affects the fidelity of final results, especially in the case of the transformation between long-sleeves and short-sleeves.
To provide structural constraints for the generation of the try-on result, we adopt Human Parsing Estimator (HPE) to predict the target parsing map, i.e., the semantic segmentation of the person wearing the target clothing.
In addition, the second role of the parsing map is to extract limb textures from the human image $I$, which is the premise of the follow-up limb-aware guidance.

Figure \ref{fig:framework} (b) illustrates the schematic of Human Parsing Estimator (HPE). Given the warped clothing $C_w$, the keypoint map $K$, and the masked source parsing map $P^s_{mask}$, this module aims to generate the target parsing map $P^t$.
The network consists of an encoder to extract the features from the original human image and a decoder to output the target parsing result. We add an additional squeeze-and-excitation (SE) block~\cite{hu2018squeeze} during downsampling to assign the weight to each channel item of the input and feature maps, which adapts its effect on the final generated image.
HPE is trained with a weighted cross-entropy loss, which is formulated as:
\begin{equation}
    L_{HPE} =-\frac{1}{n}\sum^{n}_{i=0}\sum^{c}_{j=0}{w_{j}}{P^{s}_{i,j}}log(P^{t}_{i,j})
\end{equation}
where $n$ is the sample numbers, $c$ is the number of channels of $P^s$ and $P^t$, and $w_j$ is the weight for the j-th channel.
Note that we increase the weights of the clothing and limb classes to better prevent the pixels of skin from bleeding into other regions. 

\subsection{Limb-aware Texture Fusion}
After obtaining the warped clothing $C_w$ aligned with the human body from Multi-attribute Clothing Warping and the parsing map $P^t$ of the person wearing the target clothing from Human Parsing Estimator, the final goal is to generate the try-on result by fusing textures of $C_w$ and $I$ according to $P^t$.
We observe existing methods usually generate limb regions with the wrong shape and color due to the discarding of necessary texture information of the input.
Conversely, our Limb-aware Texture Fusion (LTF) module takes full advantage of the limb textures from the human image $I$ to produce realistic limb details. Furthermore, our limb-aware guidance makes the network better handle the transformation of the clothing categories between long-sleeved and short-sleeved.
LTF consists of two stages: (1) predicting the coarse try-on result $I_{c}$ and (2) using limb map patches $L_{p}$ to refine $I_{c}$ and get the fine try-on result $I_{f}$.

\subsubsection{\textbf{Coarse Try-on}}
The first stage of LTF aims to fuse textures of the warped clothing and the human body roughly to generate the appearance of clothing regions and other regions where textures are easily transferred (e.g., face, hair, and pants). Although limb textures are also generated, there are some inaccuracies in details since the limb guidance has not been added, which are optimized in the next stage.
Given the warped clothing $C_w$, the keypoint map $K$, the target parsing map $P^t$, and the masked human image $I_{mask}$, an encoder-decoder network produces the coarse try-on result $I_c$.
Each input has its role: the warped clothing $C_w$ provides clothing textures, the keypoint map $K$ ensures the human posture is consistent with the original image, the target parsing map $P^t$ provides the structural constraints for the generated human body, and the masked human image $I_{mask}$ offers textures of other regions (e.g., face, hair, and pants) that need to be preserved.

\subsubsection{\textbf{Limb-aware Refinement}}
Experiments have demonstrated the network can utilize source limb textures to generate the result similar to the original image (as shown in Figure \ref{figure:limb} (a)).
We train our network in a self-supervised way and try to make the virtual try-on result become exactly the same as the original human image during the training process. However, it is not what we expected that this still occurs in the test set, which means that too much information from the source space causes the network to converge to a local optimum in the training set. 
This problem is especially serious when the clothing category changes during virtual try-on. For example, we try to change the short-sleeved garment in the human image to a long-sleeved one, it easily fails when the network receives the complete limb textures from the original human image, as shown in Figure \ref{figure:limb} (a), the wrong results only transfer the clothing textures but ignore the clothing geometry.
We argue the reason for this is that the rough outline of the clothing is still preserved when complete limb textures are obtained by masking the clothing along the border, which does not keep clothing-agnostic strictly. Therefore, we need to discard part of limb textures to mask the geometric appearance of the clothing.
Nonetheless, as shown in the examples in Figure \ref{figure:limb} (b) and (c), the limb textures are necessary as the auxiliary information, which effectively alleviate artifacts and distortions in the results and preserves the realistic limb details.

In order to properly utilize the guidance of the limb textures while keeping clothing-agnostic for training, we adopt the limb map patches to shield the geometric information and fuse the limb textures into the coarse try-on result $I_c$ to obtain the fine try-on result $I_f$.
Figure \ref{fig:framework} (c) illustrates the overview of the Limb-aware Texture Fusion (LTF) module.
First, the limb map $L$ is obtained by the masking operation of the target parsing map $P^t$ and the human image $I$.
Then, we divide $L$ into patches $L_p \in \mathbb{R}^{\frac{H}{s} \times \frac{W}{s}}$, where $s$ represents the patch scale.
$L_p$ items are concatenated in the channel dimension and fed into the fine try-on network with $K$, $I_{mask}$, $P^t$, and $I_c$. The fine try-on network is also a U-shaped structure with a ResNet34 and 5 decoder layers.
Note that we blur the limb regions of $I_c$ before it enters the network to narrow the gap between the training set and testing set. The blurring operation is only applied during training, which is omitted in the evaluation stage to enhance the quality of the try-on result.

\subsubsection{\textbf{Loss}}
The total loss of this module is defined as $L_{LTF}=L_{c} + L_{f}$, where $L_c$ is the loss of the coarse stage and $L_f$ is the loss of the fine stage. Each of items both contains the mean absolute error $L_1$, the perceptual loss $L_{vgg}$, and the edge loss $L_{edge}$. Note that we use $L_{edge}$ based on Sobel filters $(\bigtriangledown_x, \bigtriangledown_y)$ to correct the horizontal and vertical gradients, which improves the smoothness of the reconstructed textures. $L_c$ is defined as: 
\begin{equation}
    L_{c} = \lambda_{img}\Vert I_c - I \Vert_1 + \lambda_p\Vert\phi_i(I_c)-\phi_i(I)\Vert_1 + \lambda_{edge}\Vert\psi(I_c)-\psi(I)\Vert_1
\end{equation}
where $\phi_i(\cdot)$ represents the feature maps extracted from the $i$-th layer of the VGG19 network. $\psi(\cdot)$ denotes the gradients extracted by Sobel filters.
Similarly, $L_f$ is defined as:
\begin{equation}
    L_{f} = \lambda_{img}\Vert I_f - I \Vert_1 + \lambda_p\Vert\phi_i(I_f)-\phi_i(I)\Vert_1 + \lambda_{edge}\Vert\psi(I_f)-\psi(I)\Vert_1
\end{equation}

\section{Experiment}
In this section, we evaluate our PL-VTON on the VITON~\cite{han2018viton} dataset. 
We first introduce the VITON dataset and show the implementation details, which include the training details and the hyperparameter settings.
Then, we compare PL-VTON with the state-of-the-art methods both quantitatively and qualitatively, including CP-VTON~\cite{wang2018toward}, CP-VTON+~\cite{minar2020cp}, ACGPN~\cite{yang2020towards}, and PFAFN~\cite{ge2021parser}.
Finally, we perform ablation studies to analyze the effectiveness of each proposed contribution of PL-VTON.

\begin{figure*}[!t]
    \centering
    \includegraphics[width=1\textwidth]{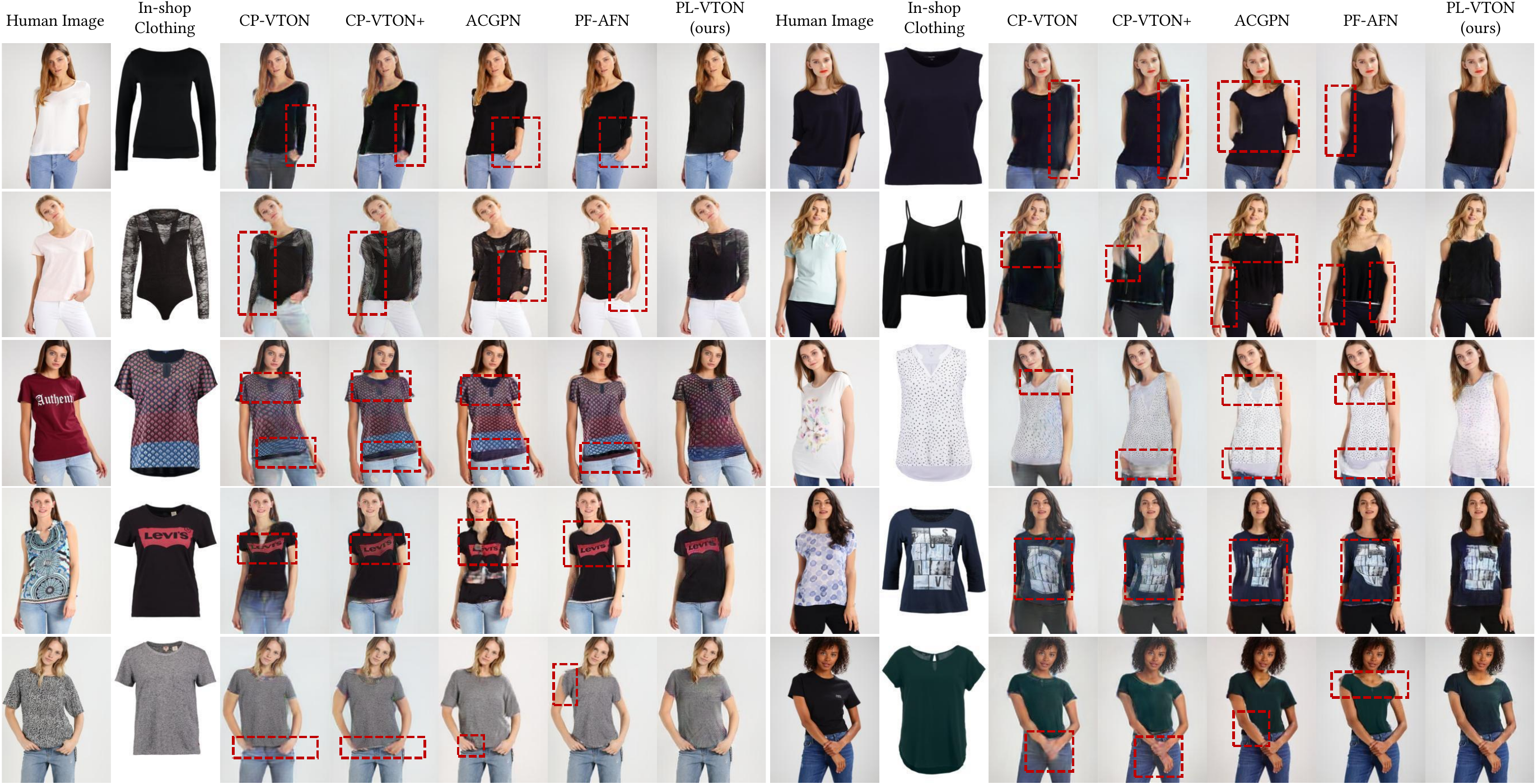}
    \caption{Visual comparisons of five different methods. PL-VTON works well for the transformation between long and short sleeves (the first row), fancy clothing try-on (the second row), cognition of the collar and hem (the third row), clothing texture transfer (the fourth row), and limb detail retention (the last row).}
    \label{fig:result} 
\end{figure*}

\subsection{Dataset}
We use the VITON~\cite{han2018viton} dataset to ensure the consistency of baseline methods. The dataset contains 19K image pairs, each pair consists of a human image and a corresponding in-shop clothing image, and the size of all images is 256 $\times$ 192. After removing the mislabeled data, there are still 16K image pairs remaining, 1/8 of which are used for testing and the others are used for training.

\subsection{Implementation Details}
Our PL-VTON is implemented with Pytorch on an RTX 3090 GPU and we train three modules independently. Specifically, we first train MCW for 24K steps with batch size 16, where the loss weights $\lambda_{mask}=2.5$, $\lambda_{cloth}=5$, $\lambda_{vgg}=1$, and $\lambda_{tv}=0.1$. Then we train HPE for 40K steps with batch size of 16, where $w_0=w_1=w_2=w_6=1$ and $w_3=w_4=w_5=3$. Finally, we train LTF for 80K steps with batch size of 4, where $\lambda_{img}=1$, $\lambda_p=2$, and $\lambda_{edge}=0.1$. Adam~\cite{kingma2014adam} optimizer is used with $\beta_1=0.5$ and $\beta_2=0.999$ in both stages. Learning rate is fixed at 0.0001 in the first half of training and then linearly decays to zero for the remaining steps.

\subsection{Qualitative Results}
We perform visual comparisons of CP-VTON~\cite{wang2018toward}, CP-VTON+~\cite{minar2020cp}, ACGPN~\cite{yang2020towards}, PF-AFN~\cite{ge2021parser}, and our PL-VTON. As shown in Figure \ref{fig:result}, where some disadvantages are highlighted with red boxes.
In the first row, we compare the quality of try-on results when the clothing category changes (i.e., the transformation between long-sleeved and short-sleeved clothing). Most existing methods do not handle the changes in the sleeve length well, and there are more artifacts in the cuff regions.
In the second row, we show the ability of methods to handle fancy clothing. Fancy clothing generally has unusual silhouettes and textures, which is extremely challenging in the virtual try-on. It can be seen that most existing methods cannot completely transfer the fancy clothing to the human body, there are some inaccurate clothing deformation and clothing texture missing in the red boxes.
\begin{table}[!t]
\centering
\setlength{\tabcolsep}{4mm}{
\caption{Quantitative evaluation results and user study results of existing methods and our PL-VTON. PL-VTON$^\ast$ is PL-VTON without the two-stage alignment strategy, and PL-VTON$^\dag$ is PL-VTON without the limb-aware guidance. For the percentage results a/b in the third column, a is the percentage that the compared method is considered better than our PL-VTON, and b is the percentage that our PL-VTON is considered better than the compared method.}
\begin{tabular}{l|ccc}
\hline
Method                          & FID             & Human                  \\  \hline
CP-VTON~\cite{wang2018toward}   & 19.88           & 21.70\% / 78.30\%      \\ 
CP-VTON+~\cite{minar2020cp}     & 16.27           & 16.67\% / 83.33\%      \\
ACGPN~\cite{yang2020towards}    & 12.69           & 24.44\% / 75.56\%      \\ 
PF-AFN~\cite{ge2021parser}      & 12.81           & 23.45\% / 76.55\%      \\   \hline
PL-VTON$^\ast$                  & 12.43           & -                      \\ 
PL-VTON$^\dag$                  & 12.58           & -                      \\   \hline
PL-VTON (ours)                  & 12.16           & reference              \\   \hline
\end{tabular}
\label{tab:metrics}}
\end{table}
In the third row, we aim to demonstrate the cognitive ability of these methods for clothing structure. Except for our PL-VTON, other methods both have some cognitive errors, especially for the collar and hem, which are easily mistaken for the front of the clothing.
The fourth row compares the ability of clothing texture transfer. Some distortions appear in the results of CP-VTON, CP-VTON+, and PF-AFN, and the textures preserved by ACPGN are incomplete.
In the last row, we examine the ability of methods to preserve limb details. For CP-VTON and CP-VTON+, there is severe blurring on the fingers, and the color tone of the generated images is different from the original. The results of ACGPN have some distortions in the finger and forearm regions. PF-AFN and PL-VTON achieve the better performance to maintain the non-clothing characteristics, but the former generates incorrect textures where the limb and clothing meet.

The above qualitative experiments show that our PL-VTON can better handle cross-category clothing transfer and fancy clothing transfer, and it is able to preserve realistic limb details while generating more accurate clothing textures.

\subsection{Quantitative Results}
We adopt Fréchet Inception Distance (FID)~\cite{heusel2017gans} to measure the feature vector distance between the real image and the generated try-on result. The higher quality of the result gets a lower score of FID. Note that Inception Score (IS) is abandoned because Rosca et.al~\cite{rosca2017variational} have demonstrated that there are misleading results when IS is applied to the model trained on the datasets other than ImageNet.
Table \ref{tab:metrics} summarizes the FID scores of CP-VTON~\cite{wang2018toward}, CP-VTON+~\cite{minar2020cp}, ACGPN~\cite{yang2020towards}, PF-AFN~\cite{ge2021parser}, and our PL-VTON on the VITON dataset. Obviously, PL-VTON achieves a significantly better FID of 12.16 compared to the second best value 12.69.

\begin{figure}[!t]
  \centering
  \resizebox{\linewidth}{!} {
    \includegraphics{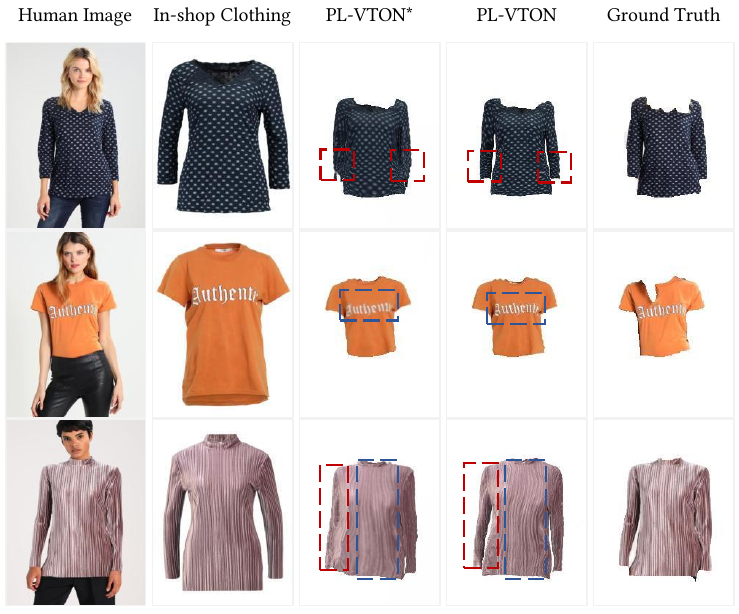}
  }
  \caption{The ablation study of the proposed two-stage alignment strategy in Multi-attribute Clothing Warping (MCW), where red boxes focus on the clothing shape and blue boxes focus on the clothing textures. PL-VTON$^\ast$ is PL-VTON without the two-stage alignment strategy.}
  \label{figure:wo_align}
\end{figure}

\subsection{User Study}
The A/B test is conducted to further evaluate the results of our method. We perform four pairwise comparisons and each comparison consists of 50 image pairs of PL-VTON and another baseline method. 50 volunteers are asked to consider the quality of these try-on images and choose the better one.
The results are summarized in Table \ref{tab:metrics}, which demonstrate that PL-VTON always gives a relatively better visual experience than other existing methods with a much higher percentage. These random and subjective tests further corroborate previous evaluation results, showing that our method significantly outperforms other methods.

\begin{figure}[!t]
  \centering
  \resizebox{1\linewidth}{!} {
    \includegraphics{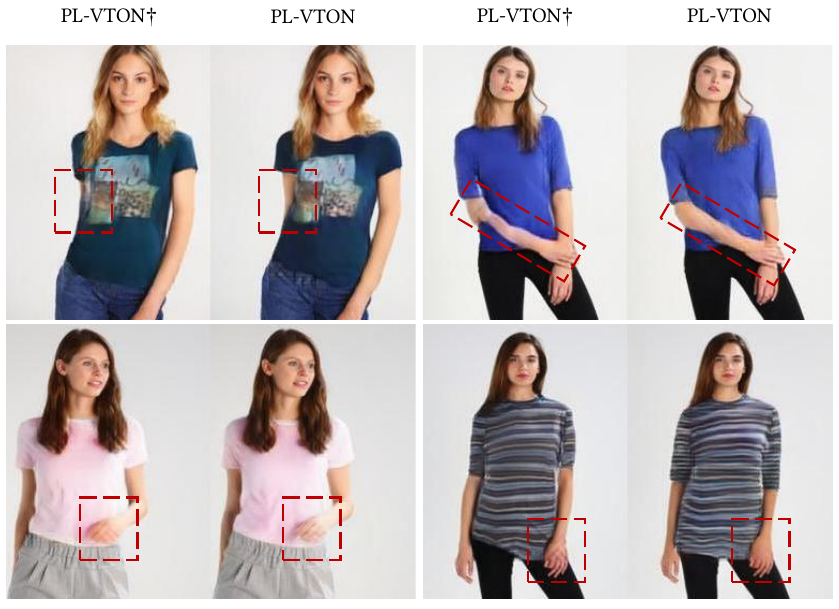}
  }
  \caption{The ablation study of the limb-aware guidance in Limb-aware Texture Fusion (LTF). PL-VTON$^\dag$ represents that the limb-aware guidance is removed from PL-VTON.}
  \label{figure:wo_limb}
\end{figure}

\subsection{Ablation Studies}
We present ablation studies on the test data to verify the role and effectiveness of the two-stage alignment strategy and the limb-aware guidance. First, we use PL-VTON$^\ast$ to indicate PL-VTON without the alignment in spatial location and size, which takes in-shop clothing $C$ as the input and directly estimates the appearance flow by a deep convolutional neural network. Then we use PL-VTON$^\dag$ to denote that the limb-aware guidance is removed from the network of Limb-aware Texture Fusion, and the virtual try-on result is produced by a single stage.
Figure \ref{figure:wo_align} shows the visual comparisons between PL-VTON$^\ast$ and PL-VTON, where red boxes focus on the clothing geometry and blue boxes focus on the clothing textures. It can be seen that compared with PL-VTON, the results of PL-VTON$^\ast$ have some wrong shape problems in the sleeve and cuff regions (the first row and the third row) and some blurred textures (the second row and the third row).
PL-VTON$^\dag$ and PL-VTON are compared in Figure \ref{figure:wo_limb}, which shows the limb regions generated by PL-VTON$^\dag$ have some distortions while the results of full PL-VTON are more realistic and accurate.
The quantitative comparisons of PL-VTON$^\ast$, PL-VTON$^\dag$, and PL-VTON are also shown in Table \ref{tab:metrics}, which indicate that the FID metric increases after removing the two-stage alignment strategy or the limb-aware guidance from PL-VTON, but the performance is still better than other methods.

\section{Conclusion}
In this paper, we propose a novel progressive limb-aware virtual try-on framework named PL-VTON. PL-VTON consists of Multi-attribute Clothing Warping (MCW), Human Parsing Estimator (HPE), and Limb-aware Texture Fusion (LTF), which produces stable clothing deformation and handles the texture retention well in the final try-on result. Extensive experiments show great superiority of PL-VTON over the state-of-the-art virtual try-on methods both qualitatively and quantitatively.

\begin{acks}
This work was supported by the National Natural Science Foundation of China (No. 61872112) and the Taishan Scholars Program of Shandong Province (No. tsqn201812106).
\end{acks}

\bibliographystyle{ACM-Reference-Format}
\bibliography{references}


\end{document}